\documentclass[letterpaper, 10pt, conference]{ieeeconf}  
\IEEEoverridecommandlockouts           
\usepackage{epsfig}			
\usepackage{amsmath}		
\usepackage{amssymb}		
\usepackage{graphicx}		
\usepackage{subfigure}		
\usepackage{url}			
\usepackage{epstopdf}		
\usepackage{listings}       

\title{\LARGE \bf A framework for Culture-aware Robots based on Fuzzy Logic}
\author{
	Barbara Bruno, Fulvio Mastrogiovanni, Federico Pecora, Antonio Sgorbissa and Alessandro Saffiotti%
	\thanks{
		B. Bruno, F. Mastrogiovanni and A. Sgorbissa are with the Department of Informatics, Bioengineering, Robotics and Systems Engineering, University of Genova,
		Via Opera Pia 13, 16145 Genova, Italy.\protect\\		
		F. Pecora and A. Saffiotti are with the AASS Cognitive Robotic Systems Lab, \"Orebro University,
		Fakultetsgatan 1, S-70182 \"Orebro, Sweden.\protect\\
		{Corresponding author's email: \tt\small barbara.bruno@unige.it}.	}%
}

\begin{document}
\begin{titlepage}  

\copyright 2017 IEEE Personal use of this material is permitted. Permission from IEEE must be obtained for all other uses, in any current or future media, including reprinting/republishing this material for advertising or promotional purposes, creating new collective works, for resale or redistribution to servers or lists, or reuse of any copyrighted component of this work in other works.

\end{titlepage}

\maketitle
\pagestyle{empty}

\begin{abstract}

Cultural adaptation, i.e., the matching of a robot's behaviours to the cultural norms and preferences of its user, is a well known key requirement for the success of any assistive application. However, culture-dependent robot behaviours are often implicitly set by designers, thus not allowing for an easy and automatic adaptation to different cultures. This paper presents a method for the design of culture-aware robots, that can automatically adapt their behaviour to conform to a given culture. We propose a mapping from cultural factors to related parameters of robot behaviours which relies on linguistic variables to encode heterogeneous cultural factors in a uniform formalism, and on fuzzy rules to encode qualitative relations among multiple variables. We illustrate the approach in two practical case studies.

\end{abstract}

\section{Introduction}
\label{sec:introduction}

In 2013, in a series of experiments involving Arab and German participants, people were asked to place a Nao robot at a suitable distance to hold a conversation with them \cite{Eresha13}. None of the participants was used to interact with a robot on a daily basis, but they all had plenty of experience in talking with other people. As a consequence, they placed the robot at a distance they deemed \emph{appropriate} for a conversation among two persons, unconsciously assuming that a robot should not be too far from a human. The experiment highlighted a difference in the behaviour of the Arab and the German participants, with the latter placing the robot farther (approx. $85$ cm) than the former (approx. $65$ cm), in accordance with the social norms of their respective \emph{cultures}.

The influence of a person's culture on his attitude towards, and preferences in the interaction with, a robot is the subject of relevant and ongoing research \cite{Carlucci15}. So far, literature findings suggest that people from different cultures not only have different preferences concerning how the robot should be and behave \cite{Evers08}, but also tend to prefer a robot that better complies with the social norms of their own culture, in aspects such as the verbal \cite{Rau09,Wang10,Andrist15} and non-verbal behaviour \cite{Trovato13} and the interpersonal distance \cite{Eresha13,Joosse14b}.

This preference does not merely affect the likeability of the robot. In a series of studies involving US, Chinese and German participants (later extended to include four other nationalities \cite{Trovato13,Andrist15}), people were asked to take decisions on an unfamiliar topic with the possibility of relying on the suggestions of a robot assistant \cite{Evers08,Rau09,Wang10}. Experimenters found that the robot's communication style has a direct impact on the number of times a person follows its advice, suggesting that people tend to trust more a robot which follows their same communication conventions.

Needless to say, to design a trustworthy robot is a key requirement for the success of any assistive application. However, culture-dependent robot behaviours are often implicitly set by designers, which makes it hard to allow for an easy and automatic adaptation to different cultures.

CARESSES\footnote{\url{www.caressesrobots.org}} is an EU-Japanese 3-year project that aims to develop and evaluate a culturally competent robot for elderly care. One of the project's goals is to make robots able to automatically adapt their behaviour to conform to a given culture; this article explores one way to achieve that goal: to devise a mapping from cultural factors to related parameters of the robot behaviours, that allows for both lateral adaptation (i.e., to switch from one person to another) and longitudinal adaptation (i.e., to follow a person's evolution over time). This goal brings about two main challenges:
\begin{enumerate}
\item culture is heterogeneous: to allow for different cultural factors, such as nationality and net income, to influence the same robot behaviour we have to find a homogeneous space for their representation; 
\item relations between cultural factors and robot behaviours are usually qualitative, and in many cases incomplete: to encode such knowledge we have to find a way to define the above mapping in qualitative, natural terms.
\end{enumerate}

We address the above challenges using techniques from fuzzy logic and fuzzy inference systems. Fuzzy logic provides us with tools to encode heterogeneous factors in a uniform formalism (linguistic variables), and to encode qualitative relations among multiple variables (fuzzy rules).  

The contribution of this paper is two-fold: (i) we propose a way to define and implement a function from variables associated with cultural factors (henceforth referred to as \textit{cultural variables}) to robot behaviour parameters, which meets the two challenges above; (ii) we describe a method for the automatic mapping of cultural variables to linguistic variables, which relies on subtractive clustering and regression. To the best of our knowledge, the work reported in this paper is the first attempt to formalize a general mapping between cultural factors and robot's behaviours.

The article is organized as follows. Section \ref{sec:problem_statement} provides background knowledge on the cultural and robotic sides, and formally defines the problem, while Section \ref{sec:method} describes the proposed solution.  Section \ref{sec:case_studies} presents two case studies which outline the capabilities of the proposed framework, and show that the results are compatible with data from the literature in
social sciences. Conclusions follow.

\section{Problem Statement}
\label{sec:problem_statement}

\begin{figure}
\centering
\includegraphics[width=8cm]{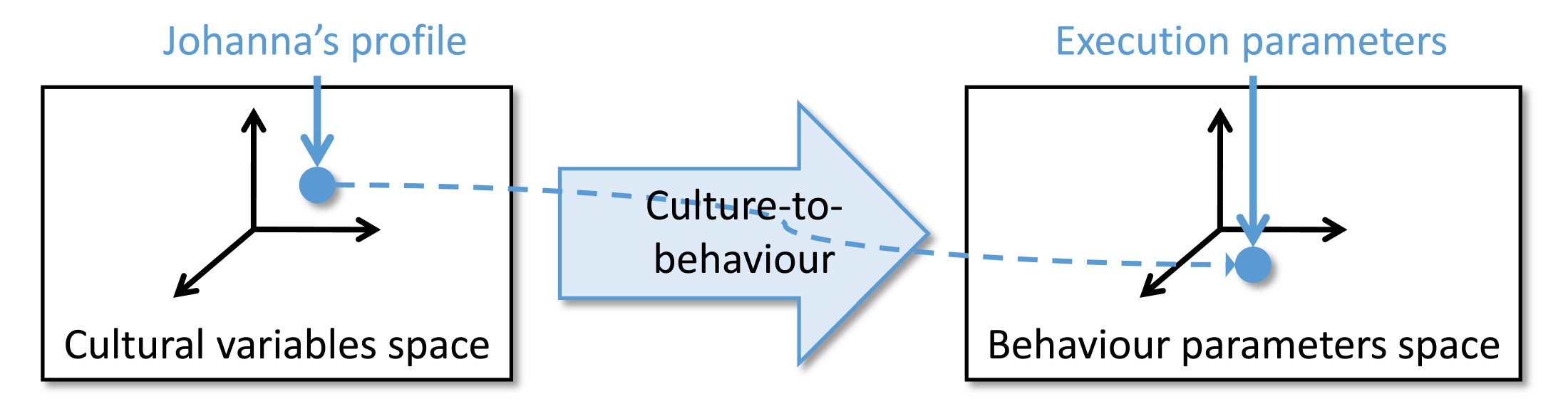}
\caption{General approach to cultural adaptation of robot's behaviour.}
\label{fig:schema}
\end{figure}

Our approach to make robots culturally aware is to use information about the cultural context to modulate the parameters that control the robot's behaviour. Figure \ref{fig:schema} illustrates this approach. The cultural context is represented by a point in the space spanned by a number of \emph{cultural variables}: in the example, the context is determined by the cultural profile of Johanna, the current user of the robot. The robot's behaviour can be modulated by a number of \emph{behaviour parameters}, like the speed of travel or the distance to keep from the human user. Our goal is to define a mapping from cultural contexts to behaviour parameters. The rest of this section further discusses the cultural variables, the robot's behaviour parameters, and the mapping between the two.

\subsection{Cultural variables}

Culture is a dynamic, multifaceted, and mostly still mysterious concept, which makes it difficult to envision a fixed and finite number of factors able to capture all its nuances. Consequently, it is impossible to list and examine all the cultural variables of relevance in a given context. However, variables can be categorized in different types, or levels of measurement, according to the nature of the information within their admissible values. The best known classification defines four
levels: nominal, ordinal, interval and ratio \cite{Stevens46}.

Nominal variables, such as gender, nationality and ethnicity, only allow for a qualitative classification. 

Ordinal variables allow for a ranking of the admissible values, while mathematical computations such as the difference are meaningless. As an example, the International Standard Classification of Education (ISCED), maintained by the UNESCO, is a framework which categorizes education in levels, ranging from 0 (pre-primary education) to 8 (doctoral or equivalent).
Hofstede's dimensions for the cultural categorization of countries are 6 scales in which the relative positions of different countries are expressed with a score on an arbitrary scale from 0 to 100 \cite{Hofstede91}. The dimension of \textit{Individualism}, for example, examines whether a nation has a preference for a loosely-knit social framework, in which individuals are expected to take care of themselves and their immediate families, or a tightly-knit one, in which individuals can expect their relatives or members of a particular in-group to look after them.

Interval variables allow for the computing the difference between items, but not the ratio, i.e., they use an equidistant scale and a non-true zero. An example of interval variable is the geographical latitude, which has two defined points (the North pole and the South pole) separated into 180 intervals.

Lastly, ratio variables possess a meaningful, unique and non-arbitrary zero value and allow for all mathematical computations. Examples include age, the number of working hours in a day, and net income.

\subsection{Robot's behaviour parameters}

\begin{figure}
\centering
\includegraphics[width=8cm]{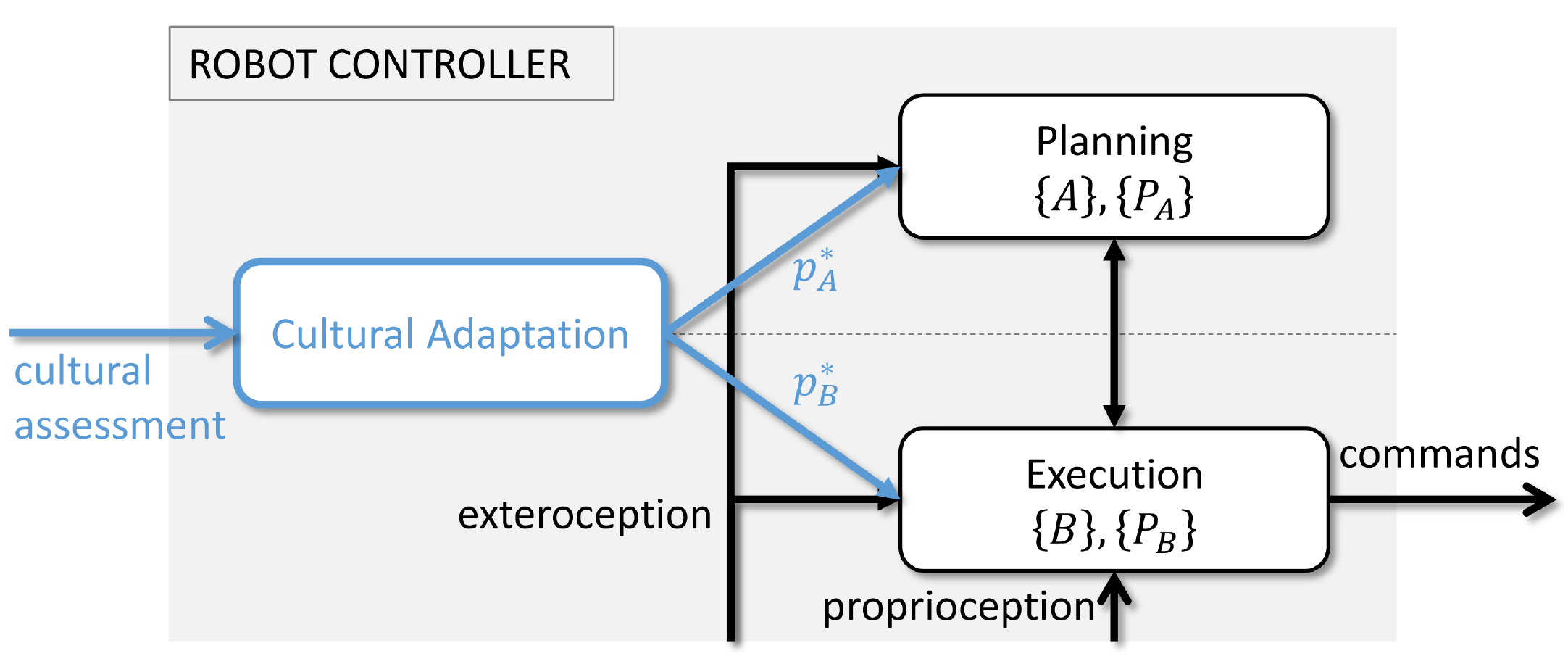}
\caption{Standard architecture of a robot high-level controller (in black) and its relation with cultural adaptation (in blue).}
\label{fig:robot_controller}
\end{figure}

Figure~\ref{fig:robot_controller} shows in black the classical Sense-Plan-Act hybrid architecture for a robot controller. In this architecture, the Planning module is responsible for generating a plan, often in the form of a sequence of robot actions, which guarantees the fulfilment of the given task while taking into account the requirements and constraints posed by the environment. To do so, the Planning module requires detailed information about the tasks it is expected to do and the actions it can perform, which constitute the planning domain and are usually specified a-priori. Each action $A$ may have parameters $P_A$, as shown in Figure~\ref{fig:robot_controller}.

Let us consider, for example, a mobile, humanoid robot for the physical assistance of elderly people. The planning domain of the robot may include actions such as \textit{Reach\_user} and \textit{Greet\_person}. The action \textit{Greet\_person} has a parameter that specifies whether its realization should be verbal, thus requiring the robot to utter a specific sentence; gestural, thus requiring the robot to wave a hand, or bow and nod with the head; or both. When planning a greeting action, the Planning module also computes the values of its parameters. Interestingly, since different types of greeting are suitable for different cultures, this decision should take into account the current cultural context.

An action typically includes several elementary behaviours, like navigating to a point, avoiding obstacles, extending an arm or grasping an object. These behaviours are executed by the Execution module, which is responsible for: (i) adapting them to the status of the robot and any change in the environment and (ii) issuing the commands for the robot's controller which will result in the desired effects. Each such behaviour $B$ may have parameters $P_B$ that modulate its execution.

For example, the robot may have a behaviour \textit{Dock\_to\_user} that takes as parameters the approach direction and the optimal distance to stop at for engaging in a conversation. Once again, the value of these parameters may depend on the preferences and cultural background of the user.

\subsection{Problem statement}

Let us denote by $\mathbb{C}$ the set of cultural variables of relevance. With no loss in generality, we may define:

\[
\mathbb{C} = \mathbb{C}_n \cup \mathbb{C}_o \cup \mathbb{C}_i \cup \mathbb{C}_r
\]

where $\mathbb{C}_n$, $\mathbb{C}_o$, $\mathbb{C}_i$ and $\mathbb{C}_r$ are the sets of cultural variables of type nominal, ordinal, interval and ratio, respectively.

Let us denote by $\mathbb{P} = \left\{P_1, P_2, \ldots, P_k\right\}$ the set of all culture-dependent robot parameters for a given application.

Given the sets $\mathbb{C}$ and $\mathbb{P}$, our problem is to define a suitable function $f: \mathbb{C} \to \mathbb{P}$ that maps values of cultural variables, possibly of different types, to values of robot's behaviours parameters. In Figure \ref{fig:schema} above, this function is represented by the ``culture-to-behaviour'' arrow.

\section{Method}
\label{sec:method}


\begin{figure}
\centering
\includegraphics[width=8cm]{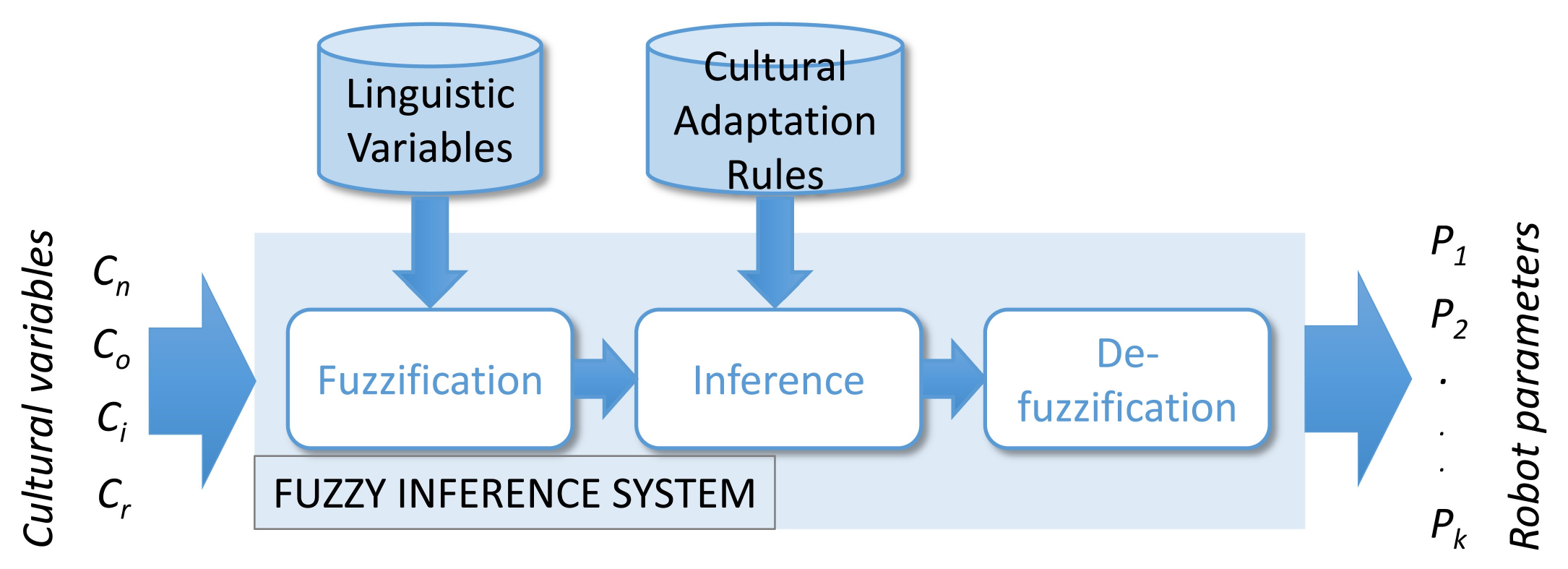}
\caption{A fuzzy inference system implementing the culture-to-behaviour function in Figure~\ref{fig:schema} above.}
\label{fig:fis}
\end{figure}

\subsection{The Fuzzy Inference System}

We define the function $f$ using a multiple-input multiple-output Fuzzy Inference System (FIS) as shown in Figure \ref{fig:fis}. The FIS implements the usual fuzzify---inference---defuzzify pipeline \cite{Driankov01}, and it uses the following knowledge: (i) a set of \emph{linguistic variables} that encode the input (cultural context) and the output (values of the robot's behaviour parameters); and (ii) a set of linguistic \emph{cultural adaptation rules} that encode how a given behaviour parameter depends on a subset of the cultural variables. For any tuple of values of the cultural variables $\vec{c} \in \mathbb{C}$, the FIS produces a tuple of values for the robot parameters $\vec{p} \in \mathbb{P}$, thus implementing the desired function $f: \mathbb{C} \to \mathbb{P}$. The shape of this function depends on the definition of the linguistic variables, and on the content of the cultural adaptation rules.

There are three main reasons why we opt for a FIS to realize our $f$ function. Two of them are technical. First, a FIS can be seen as a device to convert a linguistic specification of an input-output relation (in the form of if-then rules) into a continuous function. This feature perfectly matches our needs, since the sources in the literature that discuss the relation between cultural factors and behaviour parameters typically do so in linguistic and qualitative terms, which can be suitably encoded in the form of symbolic if-then rules. Second, linguistic variables offer a convenient mechanism to represent in a uniform way cultural variables of different types, namely, the $\mathbb{C}_n$, $\mathbb{C}_o$, $\mathbb{C}_i$ and $\mathbb{C}_r$ sets discussed above.  

The third reason is conceptual. Fuzzy logic has been given semantics in terms of similarity \cite{Ruspini.ijar1991} and desirability \cite{Ruspini.uai1991}, which fully justify its use in our work. Roughly put, suppose that the cultural adaptation rules associate a given cultural profile $\vec{c} \in \mathbb{C}$ to parameter values $\vec{p} \in \mathbb{P}$.  Then, the degree of \emph{desirability} of using $\vec{p}$ when the cultural profile is $\vec{c'} \neq \vec{c}$ is proportional to the degree of \emph{similarity} between $\vec{c'}$ and $\vec{c}$.

The key step to realize the FIS of Figure~\ref{fig:fis} is the elicitation of the linguistic variables and of the cultural adaptation rules. The remaining parts of this section discuss these issues.

\subsection{Linguistic variables}

The first bit of knowledge that we need to encode is how the input values, i.e., the values of the cultural variables, should be grouped in qualitatively meaningful \emph{linguistic variables}. In general, and following \cite{Zadeh75}, we represent a linguistic variable by the quadruple:

\[
 \langle C, \mathcal{C}, \mathcal{L}C, \mu_{LC} \rangle
\]

where $C$ is the name of the variable (e.g., \textit{Age}), $\mathcal{C}$ is its domain (e.g., $[0, 117]$ years), $\mathcal{L}C$ is the set of linguistic values $LC$ that $C$ can take (e.g., $\lbrace toddler, pre-adolescent, adolescent, adult, elderly \rbrace$) and $\mu_{LC}$ is the membership function defining the relationship between a linguistic value and the domain values.

\begin{figure}
\centering
\includegraphics[width=8cm]{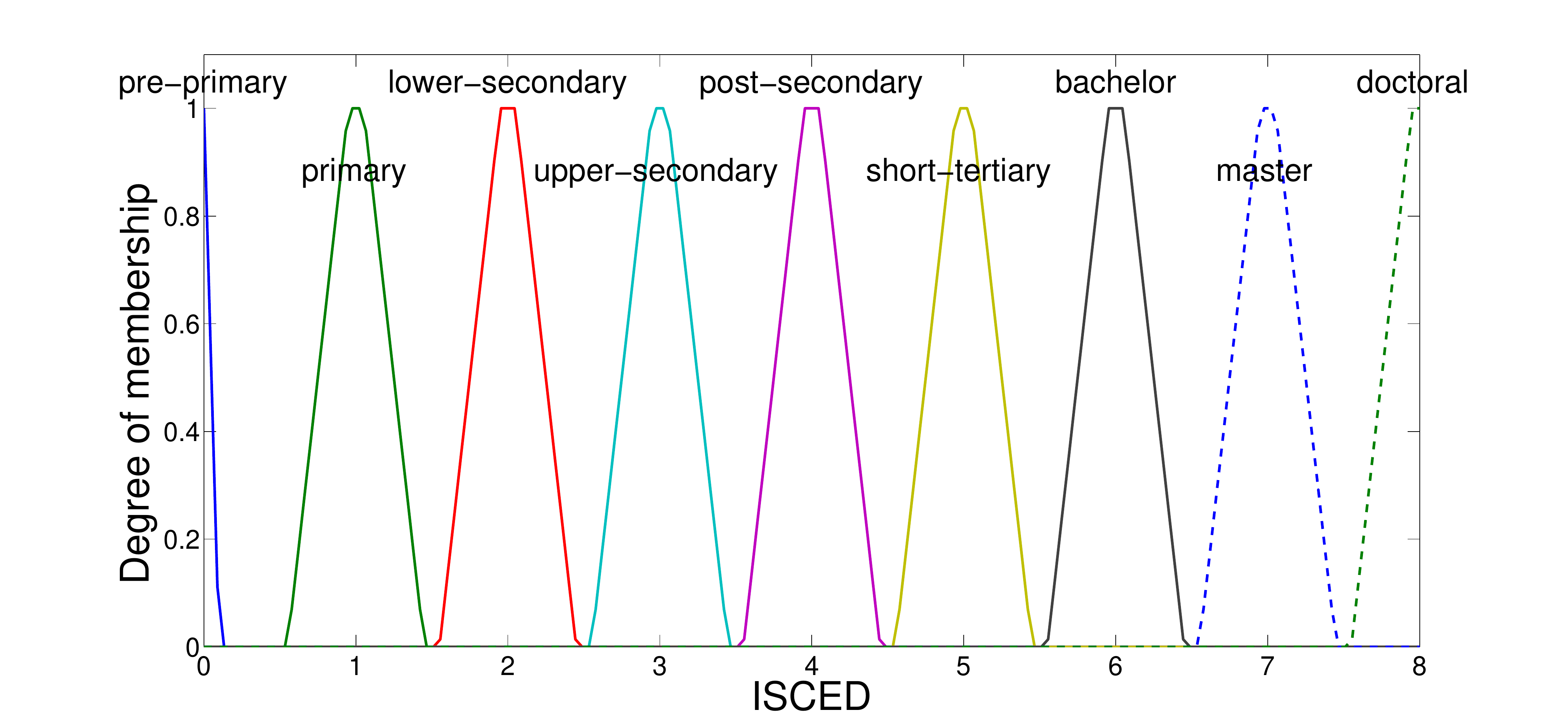}
\caption{Fuzzification of an ordinal variable of type $\mathbb{C}_o$, for which $\mathcal{C} \mapsto \mathcal{L}C$.} 
\label{fig:ISCED_example}
\end{figure}

\begin{figure}
\centering
\includegraphics[width=8cm]{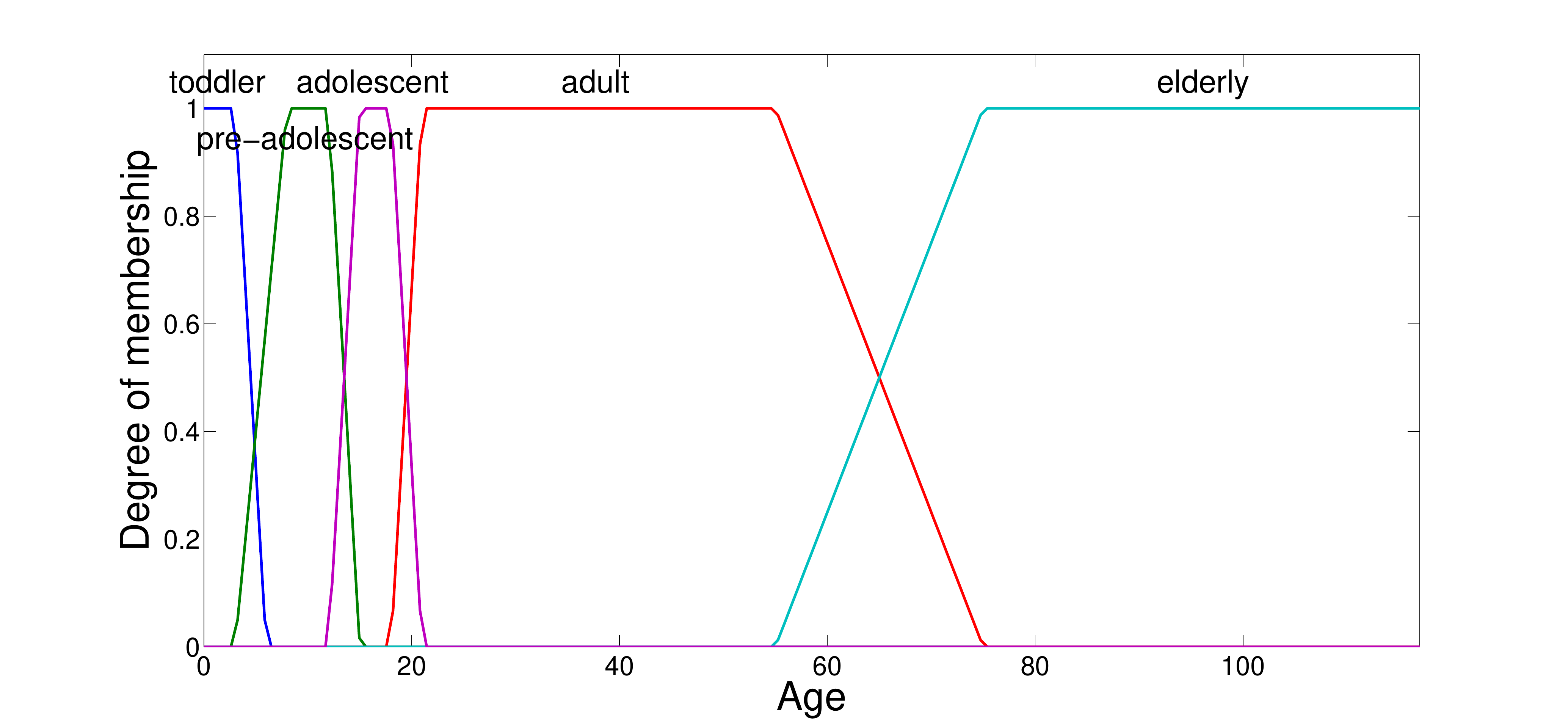}
\caption{Fuzzification of a ratio variable of type $\mathbb{C}_r$: the linguistic values correspond to the developmental stages and all membership functions are trapezoidal-shaped.} 
\label{fig:Age_example}
\end{figure}

It is possible to devise 3 cases for the mapping of a cultural variable $C$ onto a linguistic variable.

\subsubsection{$\lbrace C, \mathcal{C}, \mathcal{L}C, \mu_{LC} \rbrace$ are known}

This case is trivial, and it typically occurs with qualitative nominal or ordinal variables, for which $\vert\mathcal{C}\vert$ is finite. As an example, the linguistic variable corresponding to the \textit{ISCED} variable is shown in Figure \ref{fig:ISCED_example}.

\subsubsection{$\lbrace C, \mathcal{C}, \mathcal{L}C \rbrace$ are known, $\mu_{LC}$ are not known}

This case typically occurs with quantitative variables for which a qualitative description exists, and it requires the definition of the shape of the membership functions. As an example, the linguistic variable corresponding to the \textit{Age} variable is shown in Figure \ref{fig:Age_example}: the linguistic values are taken from the stages defined in the field of developmental psychology and the membership functions are set as trapezoidal-shaped.

\subsubsection{$\lbrace C, \mathcal{C} \rbrace$ are known, $\lbrace \mathcal{L}C, \mu_{LC} \rbrace$ are not known}

In the case of Hofstede's \textit{Individualism} dimension, for example, most studies arbitrarily impose $\mathcal{L}C = \lbrace low, medium, high \rbrace$ \cite{Rehm07,Lugrin15}, with a crisp mapping to the domain which only depends on its range. However, we argue that the introduced boundaries may be unnatural since the range is arbitrary, and propose the extraction of the linguistic values and the definition of the membership functions from available data.

Let us denote with $C^T = \lbrace c_1^T, \ldots, c_i^T, \ldots, c_I^T \rbrace$ a set of available measurements for the cultural variable of interest.

We propose a three-step procedure for the automatic estimation of $\mathcal{L}C$ and all corresponding $\mu_{LC}$ on the basis of $C^T$, based on the intuition that we can define the linguistic values $LC$ as clusters on $C^T$.

We rely on Subtractive Clustering \cite{Chiu94} for the estimation of the number of clusters to use and Fuzzy C-means Clustering \cite{Bezdek13} for the association of the points to the clusters. More specifically, the Fuzzy C-means Clustering algorithm computes for each point $c_i^T$ its membership value $\mu_{i,LC}$ to each cluster $LC$. Then, we set each membership function $\mu_{LC}$ as a two-terms Gaussian function defined as:
\[
\mu_{LC} = \alpha_1e^{-\frac{(c-\beta_1)^2}{\gamma_1^2}} +
             \alpha_2e^{-\frac{(c-\beta_2)^2}{\gamma_2^2}} 
\]
where $c$ spans the domain $\mathcal{C}$ and the parameters $\alpha_1, \beta_1, \gamma_1, \alpha_2, \beta_2, \gamma_2$ are estimated to best fit the distribution of the values $\mu_{i,LC}$. 

The result for the case of the interval variable of Hofstede's \textit{Individualism} dimension is shown in Figure \ref{fig:clustering}.

In addition to the linguistic variables representing the input, we need to encode how the possible output values of the robot's behaviour parameters should be grouped in qualitatively meaningful linguistic variables. 
This is a standard step in developing fuzzy controllers for robotic systems, and we shall not discuss it here -- see next section for an example.

\subsection{Cultural adaptation rules}

Once we have described the input and output space of our FIS in terms of fuzzy linguistic variables, the inference systems allow us to describe relations between these linguistic variables in natural language, by means of \textit{if-then} rules specified over the respective linguistic values \cite{Driankov01}. In our case, the relation to describe is the one between a given vector of values for the cultural variables, representing a given cultural context, and the corresponding robot's behaviour parameters. The \textit{if-then} rule encoding this relation can be elicited from expert knowledge: the next section provides examples of such rules. Another alternative, that we do not explore in this paper, is to learn those rules from examples.

In practice, the inference system that we use in our tests is composed of a fuzzy controller relying on Mamdani implication and composition-based inference, and on the Center-of-Area defuzzification method \cite{Driankov01}. The next section provides examples of the input-output functions realized by our FIS using the above components.

\section{Case Studies}
\label{sec:case_studies}

The validation of the proposed method requires to identify an application where cultural factors affect robot's behaviour parameters for which: (i) the relation between the cultural factors and the parameters is known, albeit in qualitative terms; (ii) there exists a dataset $C^T$ of measurements for each cultural variable for which the codomain $\mathcal{L}C$ is not known; and (iii) a number of tuples in the form of $(\vec{c}^*, \vec{p}^*)$ is available for estimating the accuracy of the inferred mapping.
 
As anticipated in the Introduction, the influence of cultural factors on the conversational distance, i.e., the distance between two persons (or a person and a robot) engaged in a conversation, is well known in the literature.
  
The first case study examines the relation between Hofstede's dimension of \textit{Individualism} and the conversational distance, and shows that our system generates results about the preferred distance that are in agreement with the literature.  The second introduces gender as a second influencing cultural variable, and shows that our system can combine the impact of multiple cultural variables on the same parameter.

Both case studies have been implemented in MATLAB (R2014a), with the Fuzzy Logic Toolbox (2.2.19) and the Curve Fitting Toolbox (3.4.1).

\subsection{From `Individualism' to `Conversational Distance'}

Let us denote by $P$ the conversational distance robot's behaviour parameter. Literature specifies that suitable values for $P$ lie in the range $\mathcal{P} = [0.45m, 1.2m]$ \cite{Joosse14b} and that this parameter is directly correlated with Hofstede's dimension of \textit{Individualism}, that we denote by $C$ \cite{Eresha13,Rehm07}. As noted above, Individualism is an ordinal variable, which expresses the relative positions of different countries with a score on an arbitrary scale $\mathcal{C} = [0, 100]$. The mapping between the two variables is only known in a qualitative form~\cite{Eresha13,Rehm07}:

\begin{quote}
countries with a high individualism score tend to have larger values for the conversational distance than countries with a low individualism score.
\end{quote}

Our goal is to define a mapping $C \to P$, induced by the above qualitative knowledge, to tune the conversational distance in accordance with the user's nationality.

\begin{figure}
\centering
\includegraphics[width=8cm]{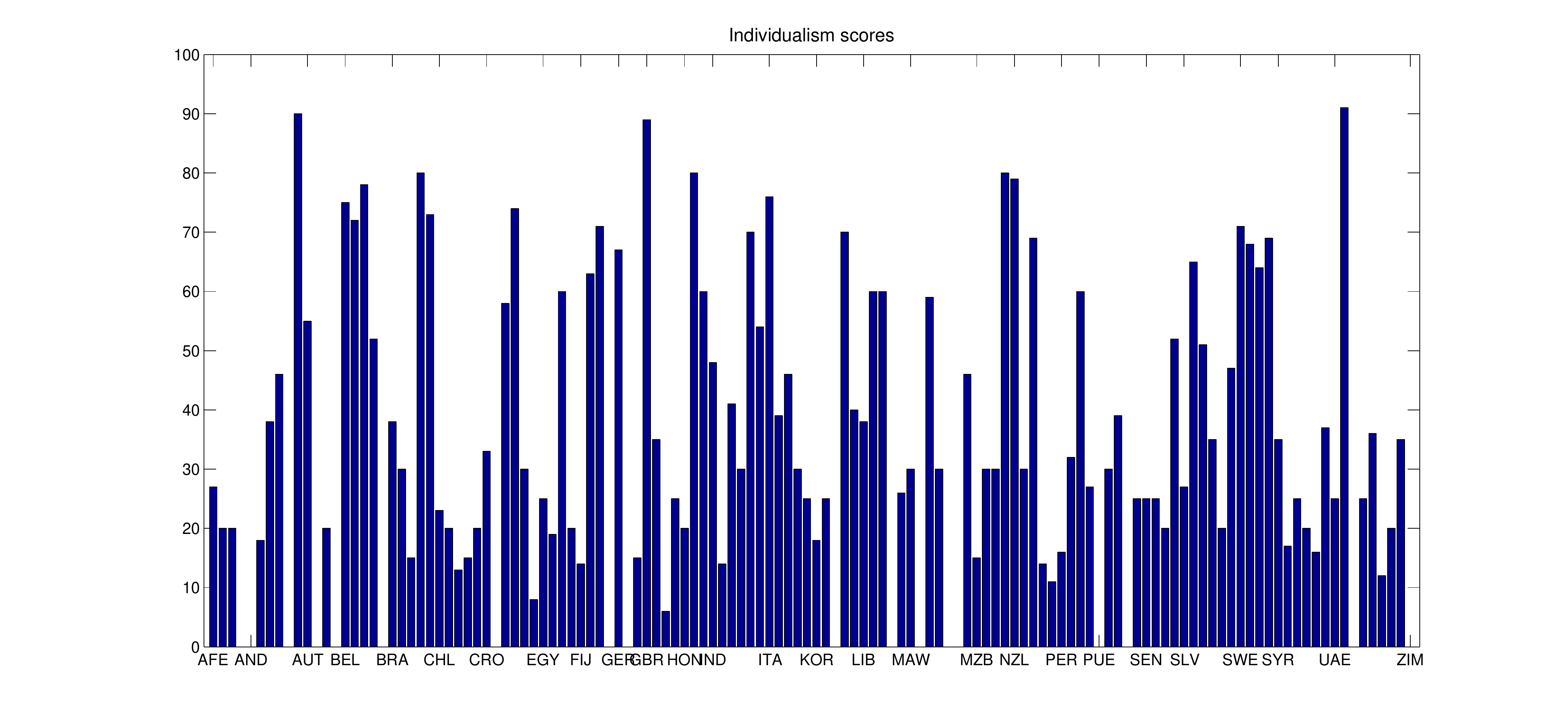}
\caption{Scores of 110 countries on Hofstede's \textit{Individualism} dimension.}
\label{fig:individualism}
\end{figure}

The first step is to define suitable linguistic variables for the \textit{Individualism} dimension. We follow the approach outlined in Section \ref{sec:method} to compute an \text{a-posteriori} partition based on the analysis of the existing data.

\begin{figure}
\centering
\includegraphics[width=8cm]{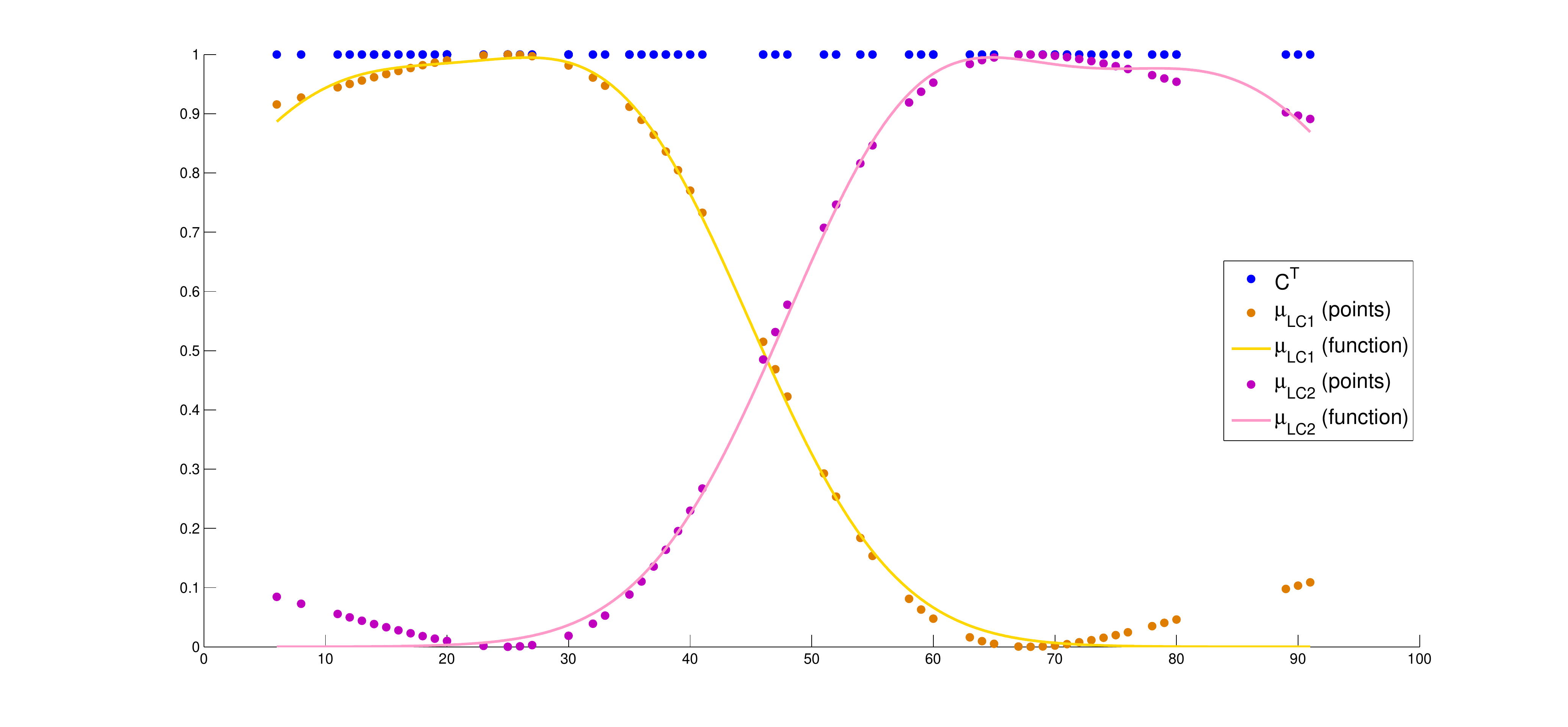}
\caption{Clustering \& Fuzzification of the training set $C^T$ of the \textit{Individualism} cultural variable $C$.} 
\label{fig:clustering}
\end{figure}

Literature provides the \textit{Individualism} scores of $110$ countries\footnote{The scores are publicly available at: \url{http://geerthofstede.com/research-and-vsm/dimension-data-matrix/}}, shown in Figure \ref{fig:individualism}. We use these scores as training dataset $C^T$ for the \textit{Individualism} variable. The x-axis of Figure \ref{fig:clustering} spans the domain $\mathcal{C}$ and the blue dots mark the $110$ scores (e.g., $c^T_1 = 6$ corresponds to the score of Guatemala and $c^T_2 = 8$ corresponds to the score of Ecuador).

In accordance with the method outlined in Section \ref{sec:method}: (i) the Subtractive Clustering algorithm identifies $2$ as the optimal number of clusters; (ii) Fuzzy C-means Clustering computes the membership values $\mu_{i,LC1}$ to cluster $LC1$ (orange dots) and the membership values $\mu_{i,LC2}$ to cluster $LC2$ (purple dots). Finally, each membership function $\mu_{LC}$ is computed as the two-terms Gaussian function best fitting the distribution of the values $\mu_{i,LC}$. In Figure~\ref{fig:clustering}, the yellow line corresponds to $\mu_{LC1}$ and the pink line to $\mu_{LC2}$.

\begin{figure}
\centering
\includegraphics[width=8cm]{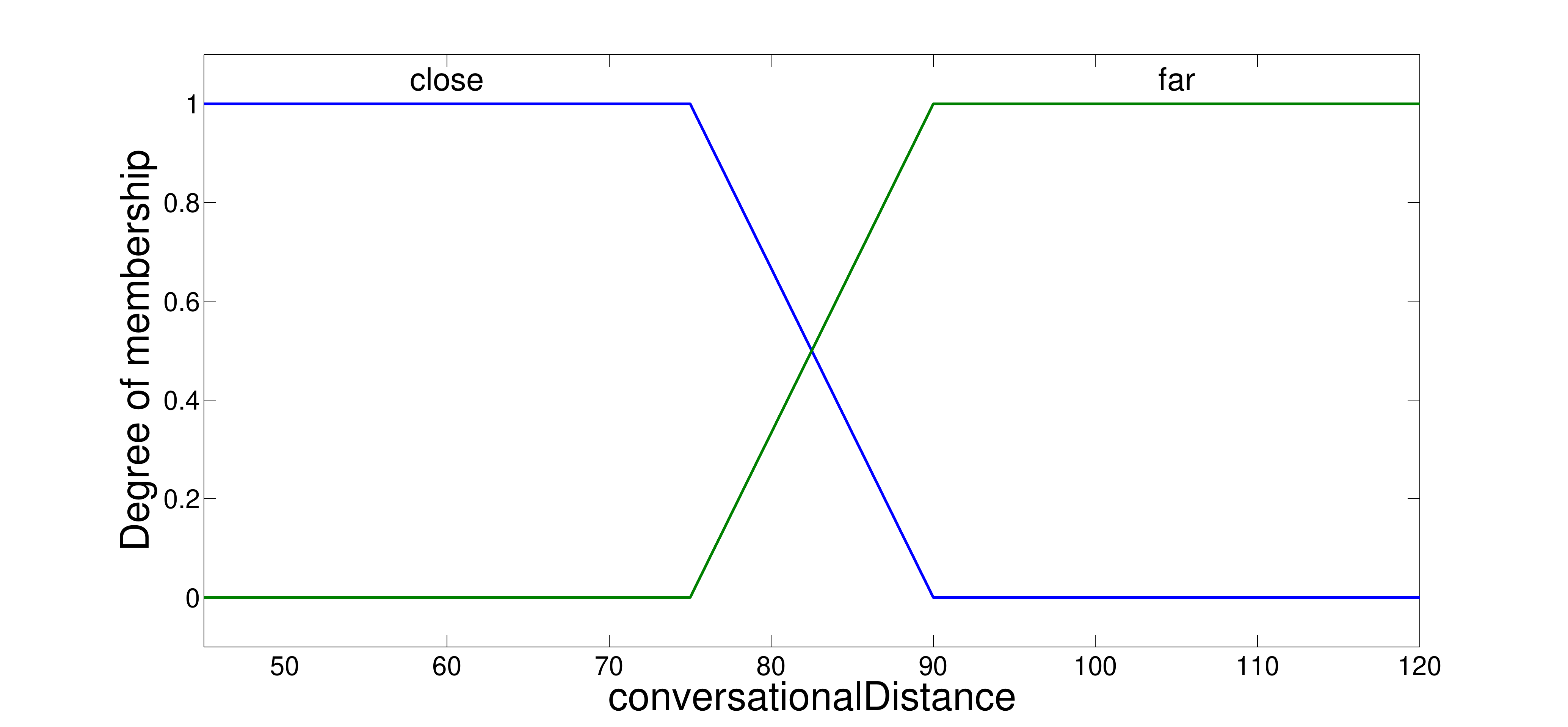}
\caption{Description of the \textit{Conversational distance} as a linguistic variable.}
\label{fig:conversationalDistance}
\end{figure}

The second linguistic variable that we need is the one representing the robot behaviour parameter $P$. For this, we simply define the linguistic variable $\langle P, \mathcal{P}, \mathcal{L}P, \mu_{LP} \rangle$, with $\mathcal{L}P = \lbrace close, far \rbrace$ as shown in Figure \ref{fig:conversationalDistance}.  

We are now in a position to encode the qualitative knowledge about the relation between the cultural variable and the parameter variable. We do so using the following rules:

{\footnotesize
\begin{equation}
\left\lbrace
\begin{aligned}
&\texttt{if } C \texttt{ is } LC1 \texttt{ then } P \texttt{ is } close \\
&\texttt{if } C \texttt{ is } LC2 \texttt{ then } P \texttt{ is } far
\end{aligned}
\right.
\label{eq:rules}
\end{equation}
}

\begin{figure}
\centering
\includegraphics[width=8cm]{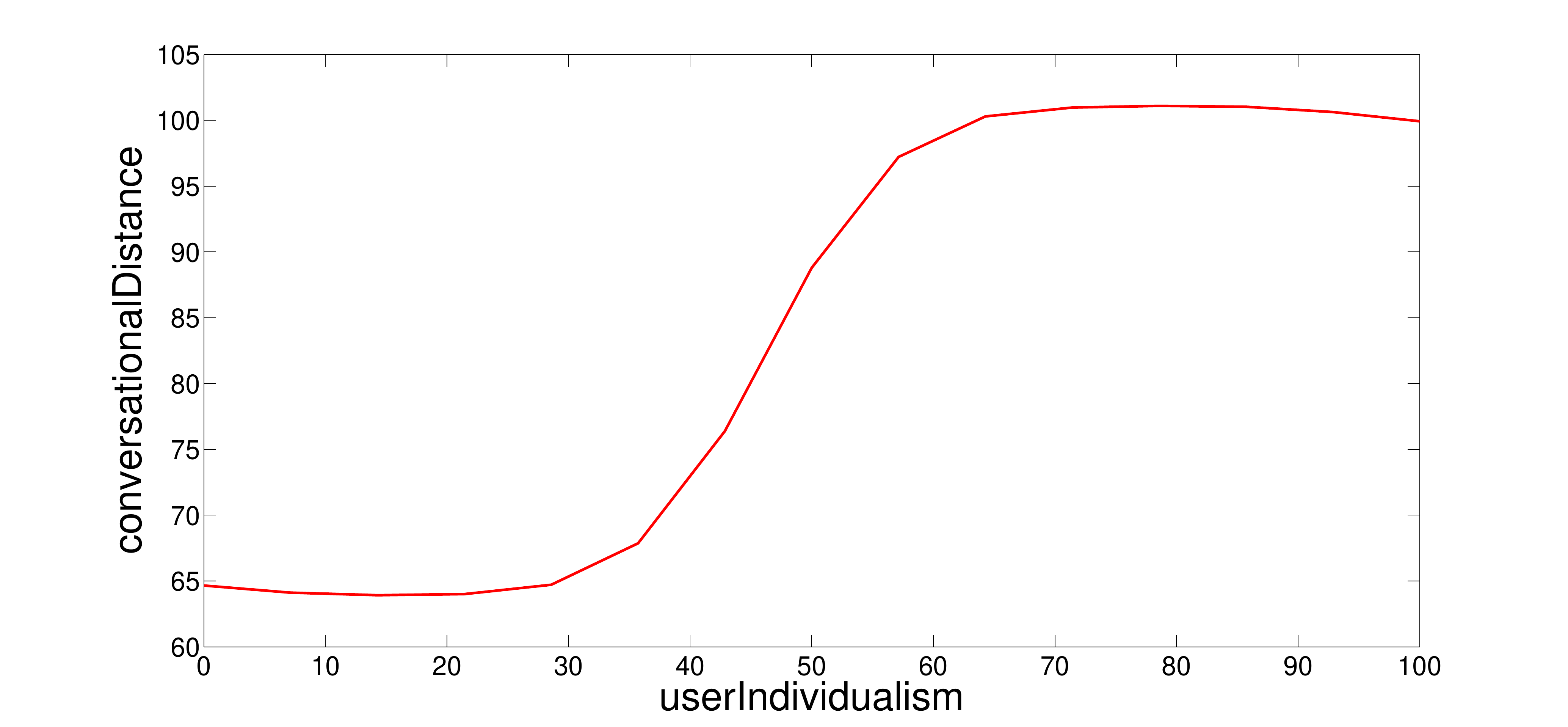}
\caption{\textit{Individualism} $\to$ \textit{Conversational distance} given by \eqref{eq:rules}.} 
\label{fig:fuzzyController}
\end{figure}

Using the linguistic variables above together with the set of rules specified in~\eqref{eq:rules}, our FIS generates the desired continuous mapping $\mathcal{C} \to \mathcal{P}$, which is plotted in Figure \ref{fig:fuzzyController}. 

Below are two examples of the values for the conversational distance generated from two different values of the \textit{Individualism} cultural factor, corresponding to the average scores of Arab countries and of Germany, respectively. These results agree with the experimental results in \cite{Eresha13}.

{\footnotesize
\[
\begin{aligned}
&c = 38 \rightarrow p = 69.9cm \\ 
&c = 67 \rightarrow p = 100.7cm   
\end{aligned}
\]
}

\subsection{From `Individualism' and `Gender' to `Conversational Distance'}

In addition to Hofstede's \textit{Individualism} dimension, \textit{gender} is another factor known to influence the conversational distance \cite{Sussman82}. The mapping between the two cultural variables and the robot parameter is again only qualitative:

\begin{quote}
countries with a high individualism score tend to have larger values for the conversational distance than countries with a low individualism score, and males tend to have larger values for the conversational distance than females.
\end{quote}

\begin{figure}
\centering
\includegraphics[width=8cm]{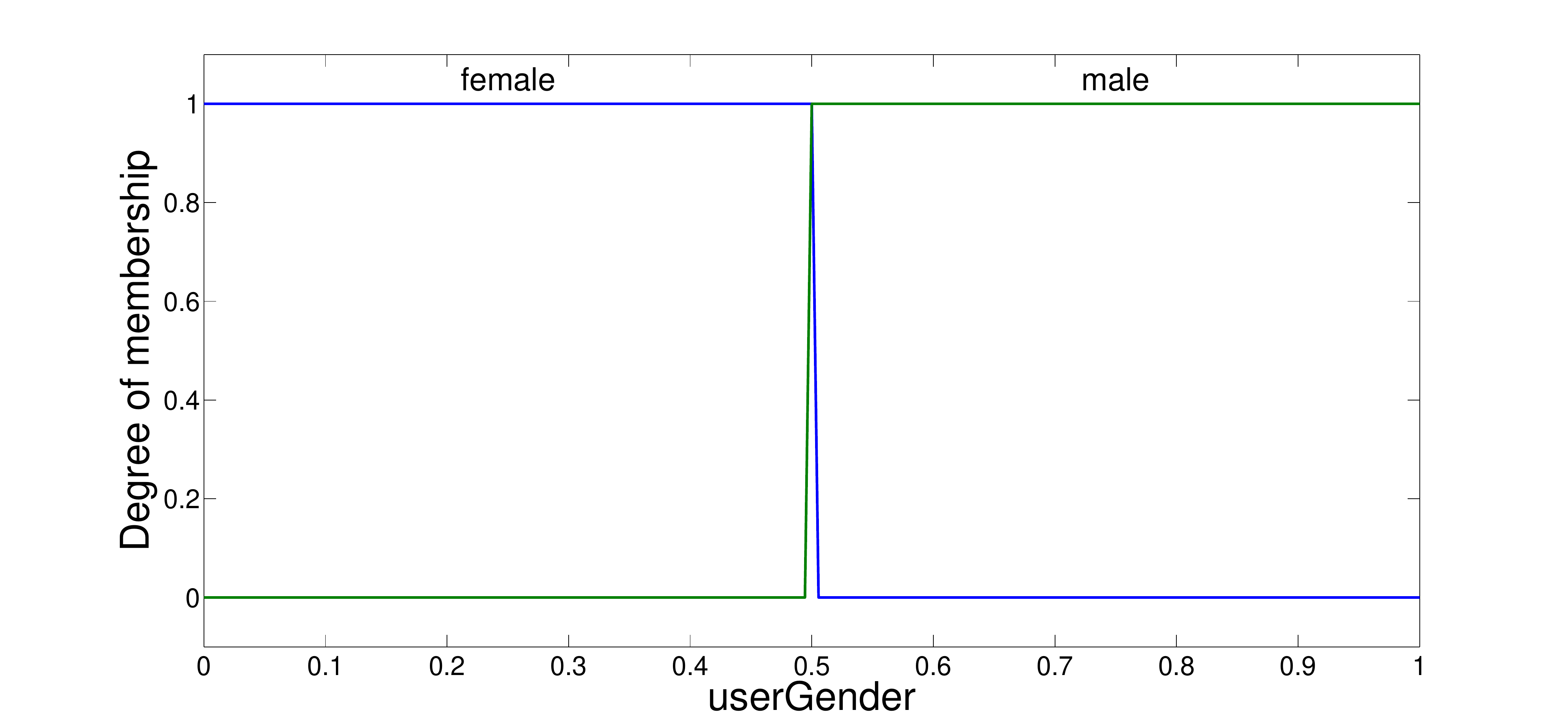}
\caption{Fuzzification of the nominal variable \textit{Gender} (case 2): membership functions have been arbitrarily set as trapezoid-shaped.} 
\label{fig:Gender_example}
\end{figure}

The fuzzification of \textit{Gender}, that we denote with $C_2$, falls into the second of the three cases examined in Section \ref{sec:method}: the corresponding linguistic variable is shown in Figure \ref{fig:Gender_example}.

\begin{figure}
\centering
\includegraphics[width=8cm]{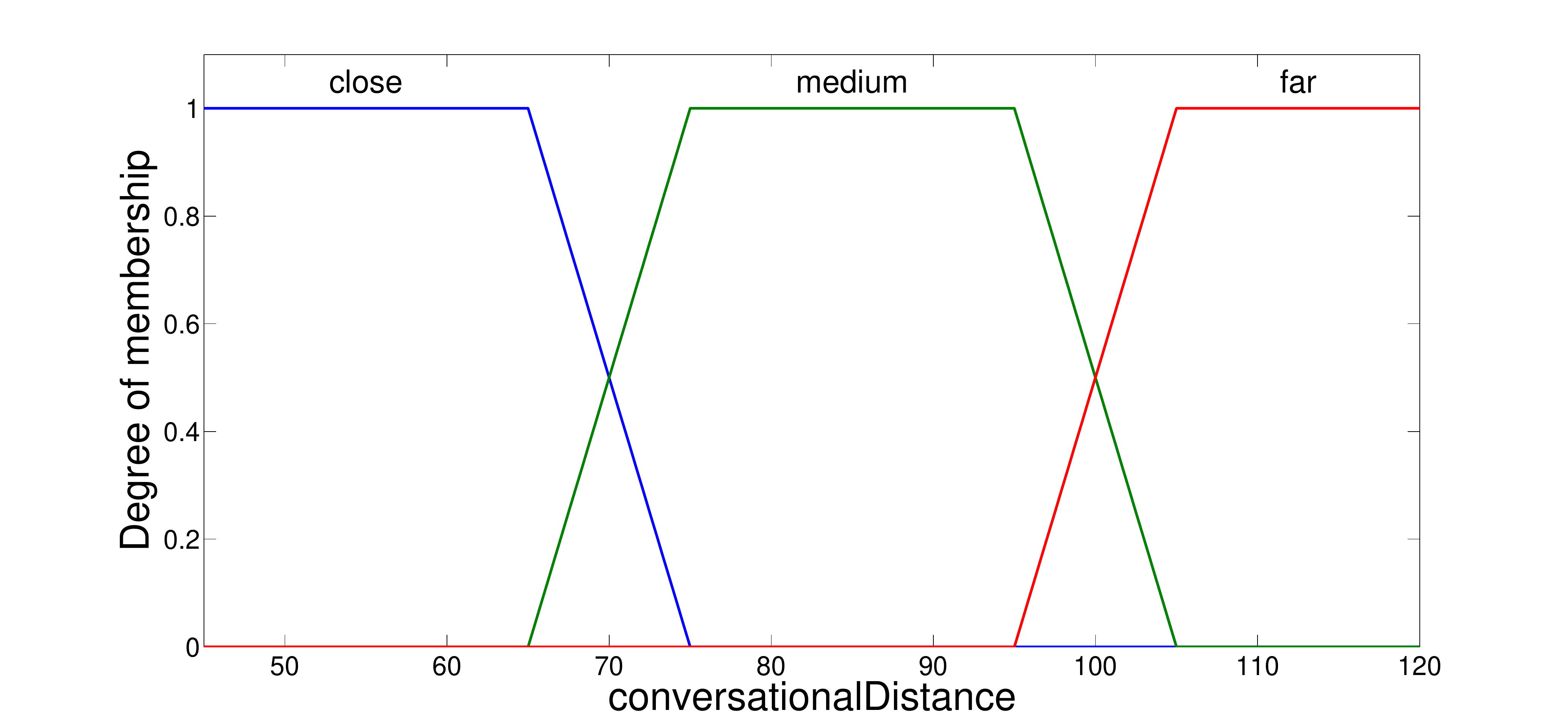}
\caption{Description of \textit{Conversational distance} as a linguistic variable, with $\mathcal{L}P = \lbrace close, medium, far \rbrace$.}
\label{fig:conversationalDistance_2}
\end{figure}

\begin{figure}
\centering
\includegraphics[width=8cm]{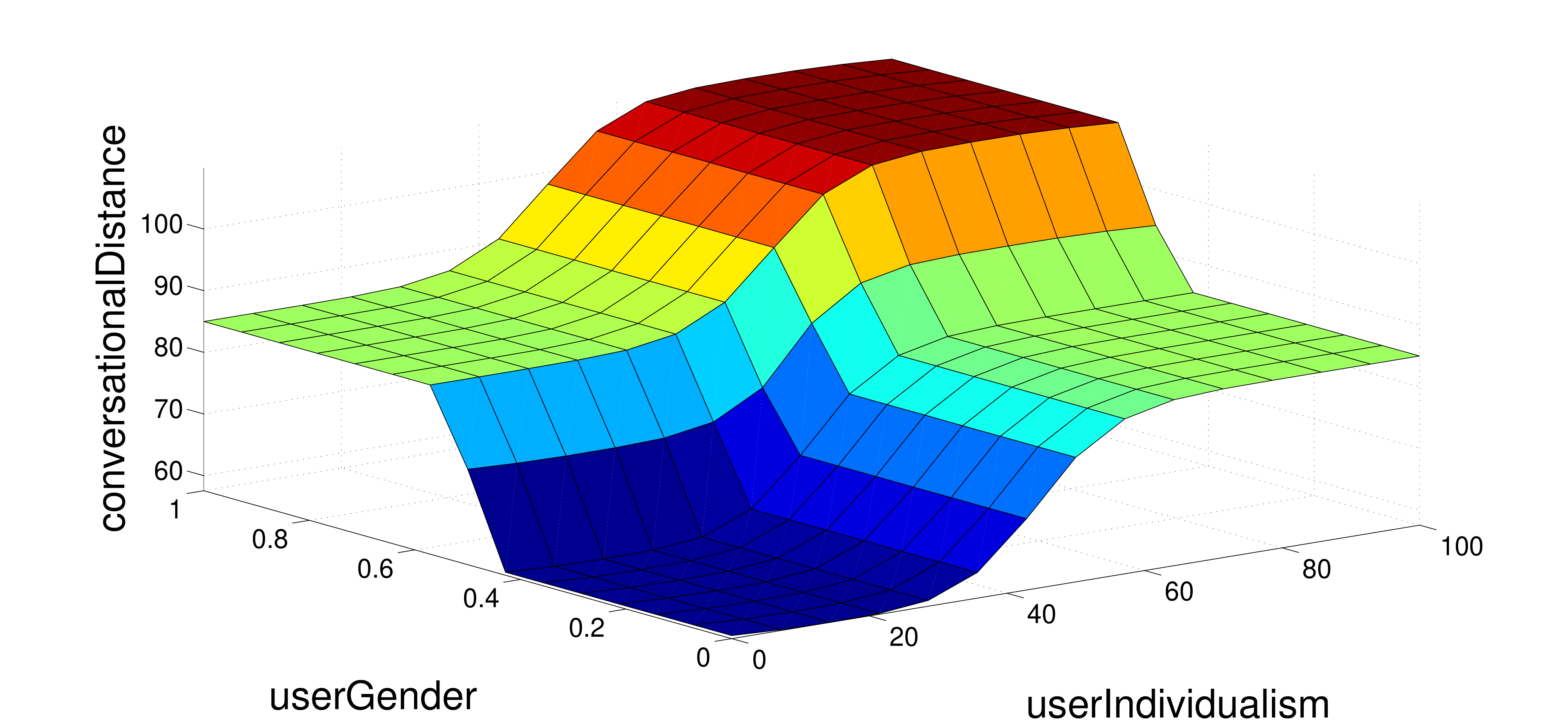}
\caption{\textit{Individualism}, \textit{Gender} $\to$ \textit{Conversational distance} given by \eqref{eq:rules_2}.}
\label{fig:fuzzyController_2}
\end{figure}

We encode the above qualitative knowledge in the rules:

{\footnotesize
\begin{equation}
\left\lbrace
\begin{aligned}
&\texttt{if } C \texttt{ is } LC1 \texttt{ and } C_2 \texttt{ is } female \texttt{ then } P \texttt{ is } close \\
&\texttt{if } C \texttt{ is } LC2 \texttt{ and } C_2 \texttt{ is } female \texttt{ then } P \texttt{ is } medium \\
&\texttt{if } C \texttt{ is } LC1 \texttt{ and } C_2 \texttt{ is } male \texttt{ then } P \texttt{ is } medium \\
&\texttt{if } C \texttt{ is } LC2 \texttt{ and } C_2 \texttt{ is } male \texttt{ then } P \texttt{ is } far
\end{aligned}
\right.
\label{eq:rules_2}
\end{equation}
}
and refine the linguistic variable associated with parameter \textit{P} as shown in Figure \ref{fig:conversationalDistance_2}. The linguistic variables $C, C_2, P$ together with the set of rules specified in~\eqref{eq:rules_2} generates the non-linear mapping $\lbrace \mathcal{C}, \mathcal{C}_2 \rbrace \rightarrow \mathcal{P}$ shown in Figure \ref{fig:fuzzyController_2}. Notice that the resulting mapping does not correspond to a simple composition of independent mappings of the input variables onto the output. As an example:

{\footnotesize
\[
\begin{aligned}
&\vec{c} = [38,0] \rightarrow p = 63.63cm \quad (Arab\_woman)\\
&\vec{c} = [67,0] \rightarrow p = 84.7cm \quad (German\_woman)\\
&\vec{c} = [38,1] \rightarrow p = 87.51cm \quad (Arab\_man)\\
&\vec{c} = [67,1] \rightarrow p = 109.34cm \quad (German\_man).
\end{aligned}
\]
}

\section{Conclusions}
\label{sec:conclusions}

The problem of designing culture-aware robots, able to adapt their behaviour to conform to a given culture, has emerged only very recently in the field of autonomous robotics. The importance of this problem, however, will most probably raise dramatically as robots enter our homes and working spaces. The approach presented here is a first step in building the capability to be culture-aware: we are not aware of previous attempts in this direction. We speculate that our approach can be used beyond robotics, to add culture-aware capabilities to generic computer systems that interact with humans.
The approach has been illustrated on two simple cases, and we have shown that it can produce results in agreement with the literature in social psychology.

While these results are encouraging, more and larger scale experiments are needed to fully validate our approach, where the collaboration with experts in cultural diversity will be essential to elicit the relevant knowledge. Work in this direction is currently undergoing in the framework of the collaborative EU-Japan project CARESSES.


\section{Acknowledgement}
 
This work has been partly supported by a grant of the Fondazione/Stiftelsen C.M. Lerici awarded to the first author; by the Italian Ministry of Foreign Affairs and International Cooperation (MAECI) and the Italian Ministry of Education, Universities and Research (MIUR) under grant No. PGR00193 (WEARAMI) and by the European Commission Horizon2020 Research and Innovation Programme under grant agreement No.~737858 (CARESSES).

\bibliography{bibliografia}

\end{document}